\documentclass[conference]{IEEEtran}
\IEEEoverridecommandlockouts
\usepackage{cite}
\usepackage{amsmath,amssymb,amsfonts}
\usepackage{algorithmic}
\usepackage{graphicx}
\usepackage{textcomp}
\usepackage{xcolor}

\usepackage{microtype}
\usepackage[justification=centering]{caption}

\usepackage[noend,linesnumbered,ruled]{algorithm2e}

\usepackage{todonotes}
\usepackage{multirow}

\newcommand{\cmaes}{CMAES}
\newcommand{\sepcmaes}{SepCMA}
\newcommand{\snes}{SNES}
\newcommand{\xnes}{XNES}

\newcommand{\halfcheetah}{\textsc{HalfCheetah}}
\newcommand{\walker}{\textsc{Walker2D}}
\newcommand{\hopper}{\textsc{Hopper}}

\newcommand{\swimmer}{\textsc{Swimmer}}

\usepackage{color, colortbl}
\usepackage{url} 
\usepackage{amsthm}
\usepackage{amssymb,amsmath,bbm}
\usepackage{dsfont}
\usepackage{array}
\usepackage{bm}
\usepackage{multirow}
\usepackage[footnote]{acronym}

\begin{document}

\title{Searching Search Spaces: Meta-evolving a Geometric Encoding for Neural Networks}

\author{
    \IEEEauthorblockN{Tarek Kunze}
    \IEEEauthorblockA{
        \textit{ISAE-SUPAERO,}\\
        \textit{Université de Toulouse,}\\
    Toulouse, France \\
    tarek.kunze@protonmail.com}
\and
    \IEEEauthorblockN{Paul Templier}
    \IEEEauthorblockA{
        \textit{ISAE-SUPAERO,}\\
        \textit{Université de Toulouse,}\\
    Toulouse, France \\
    paul.templier@isae.fr}
\and
    \IEEEauthorblockN{Dennis G. Wilson}
    \IEEEauthorblockA{
        \textit{ISAE-SUPAERO,}\\
        \textit{Université de Toulouse,}\\
    Toulouse, France \\
    dennis.wilson@isae.fr}
    
}

\maketitle

\begin{abstract}
    In evolutionary policy search, neural networks are usually represented using a direct mapping: each gene encodes one network weight. Indirect encoding methods, where each gene can encode for multiple weights, shorten the genome to reduce the dimensions of the search space and better exploit permutations and symmetries. 
    The Geometric Encoding for Neural network Evolution (GENE) introduced an indirect encoding where the weight of a connection is computed as the (pseudo-)distance between the two linked neurons, leading to a genome size growing linearly with the number of genes instead of quadratically in direct encoding. However GENE still relies on hand-crafted distance functions with no prior optimization. Here we show that better performing distance functions can be found for GENE using Cartesian Genetic Programming (CGP) in a meta-evolution approach, hence optimizing the encoding to create a search space that is easier to exploit. We show that GENE with a learned function can outperform both direct encoding and the hand-crafted distances, generalizing on unseen problems, and we study how the encoding impacts neural network properties. 

\end{abstract}

\begin{IEEEkeywords}
    evolution strategies, genetic programming, meta-evolution, encoding, neural networks, reinforcement learning, policy search
\end{IEEEkeywords}

\section{Introduction}
\label{sec:intro}

Artificial Neural Networks (ANNs) are at the heart of modern deep learning systems, behind applications in computer vision, control, and language processing \cite{goodfellow2016deep}. The Universal Approximation Theorem \cite{cybenko_approximation_1989} posits that any function can be approximated by a neural network, but efficiently finding the architecture and parameters that define the neural network approximating a function has been the focus of research for the last three decades. 

In many cases where gradients are hard to obtain, Evolutionary Algorithms (EA) have been used to optimize neural networks. Policy search, e.g. finding the network that represents the best decision function ("policy") for an agent interacting with an environment, presents a situation where the optimal policy is not known in advance hence the exact error cannot be computed. EAs have been successfully applied to policy search, yielding results competitive with gradient-based Reinforcement Learning (RL) approaches \cite{salimans_evolution_2017}. Using an EA to optimize the parameters of an ANN requires representing the network as a genome: the most straightforward  approach is to concatenate its parameters in one vector. In this "direct" encoding each gene represents a single parameter. As most neural network architectures rely on fully-connected layers where each neuron from a layer takes as inputs the activation values of all the neurons from the previous layer, the number of weights in one of these layers grows quadratically with the number of neurons and so does the size of the genome which in turns increases the optimization cost \cite{templier_geometric_2021}. 

To reduce the genome size scaling to linear with the number of neurons, the Geometric Encoding for Neural network Evolution (GENE, \cite{templier_geometric_2021}) framework introduced an indirect encoding method where the weight of each connection is generated by computing the (pseudo-)distance between the coordinates of the two connected neurons in a latent space. The neuron coordinates are then optimized with the EA, each coordinate being used to generate the weights of all incoming and outgoing connections for that neuron. Smaller genomes reduced the optimization time and memory costs by 2 orders of magnitude with the Exponential Natural Evolution Strategy (\xnes{}, \cite{wierstra_natural_2008}) on Atari control tasks. 
However GENE relies on hand-crafted distance functions and is still outperformed by direct encoding on many tasks, showing the subspace of neural networks attainable through the encoding is either harder to search or does not allow to represent best-performing policies that direct encoding would allow. In this work we propose to optimize the distance function with a meta-evolution process, using Cartesian Genetic Programming (CGP, \cite{miller_cartesian_2008}) to represent it as an explainable graph. We show that CGP can find a simple GENE encoding function that outperforms the hand-crafted method and direct encoding, generalizing to control problems not seen during training.

\section{Related works}
\label{sec:related}

Optimizing a neural network usually combines two parts: defining its architecture by choosing which neurons are connected, and fixing the value of the weight on each connection. These steps can be done at the same time with methods such as Neuroevolution of Augmenting Topologies (NEAT, \cite{stanley_efficient_2002}) which builds the network by progressively adding connections, or separately where an outer loop searches for the architecture while an inner method optimizes the parameters of a fixed architecture. While the architecture optimization step is studied by the field of Neural Architecture Search \cite{NAS_Survey}, we focus on the task of finding the optimal parameters for a fixed architecture.  

\subsection{Encoding neural networks}
Using neuroevolution as an optimization methods for deep neural networks requires the parameters of the network to be encoded as a genome.
Direct encoding concatenates all the weights and biases of a neural network into one vector $\sigma \in \mathbb{R}^D$, the genome. The size of the genome grows linearly with the number of connections in the network. However when connecting two consecutive layers of size $a$ and $b$, there are $a \times b$ weights to consider. The size of the genome then grows quadratically with the number of neurons in each layer.

Inspired by the biological genotype-to-phenotype mapping where one gene can impact the development of many parts of the body, indirect encoding methods aim to change the search space by adding a transformation between the genome and the neural network it defines. This transformation can reduce the dimensions of the genome reducing optimization costs, but also change the typology of the search space to help find high-performing solutions as neural networks show many permutations and symmetries which could be exploited. As in this work we focus specifically on the GENE indirect encoding \cite{templier_geometric_2021} (see section \ref{sec:GENE}), we refer to their paper for a more detailed review of indirect encoding methods.

\subsection{Optimizing neural networks}

Once the encoding of policy networks into genomes as vectors of continuous values is defined, policies can be optimized as black-box problems with evolutionary methods such as genetic algorithms \cite{mitchell_introduction_1998} or evolution strategies (ES, \cite{rechenberg1978evolutionsstrategien}). The evaluation of a genome is done by building the network with the weights from the genome, and using the network as a policy in an episode of the problem. The total reward over the episode is then used as fitness for the evolutionary algorithm. Optimizing the parameters of the genome is hence equivalent to optimizing the policy the network represents. 
Usually used for the continuous optimization of parameters in a neural network, Evolution Strategies are a subset of evolutionary algorithms that rely on a distribution in the genome space to sample solutions to evaluate. The distribution is then updated based on the ranking of solutions to favor the best performing ones, iteratively until convergence. ES have been shown to be competitive with deep reinforcement learning methods on robotics and image-based control tasks \cite{salimans_evolution_2017}. 

In the evolution of neural networks, the ES is not aware of the encoding used since the process is black-box. The encoding however changes the search space by potentially reducing its dimensionality and how it can be navigated. While characterizing the optimization landscape is outside the scope of this work, the change in genome size can have a direct impact on optimization cost. This is especially the case with ES methods that compute the covariance matrix of the distribution like Covariance Matrix Adaptation Evolution Strategy (\cmaes{}, \cite{hansen_cma_2016}) or Exponential Natural Evolution Strategy (\xnes{}, \cite{wierstra_natural_2008}). As their compute cost grows quadratically with the size of the genome, their use is still limited to small policies. Variants of these methods like Separable \cmaes{} (\sepcmaes{}, \cite{ros2008simple}) or Separable Natural Evolution Strategy (\snes{}, \cite{wierstra_natural_2008}) have been introduced to tackle the cost by considering all optimization dimensions as separable, which reduces the compute cost but impacts performance.

\subsection{A Geometric Encoding}
\label{sec:GENE}

The Geometric Encoding for Neural Network Evolution (GENE, \cite{templier_geometric_2021}) projects all the neurons of a fully-connected neural network into a latent Euclidean space of small dimensions $d$, then optimizes the concatenated coordinates of these neurons using Evolution Strategies. To generate the network used as a policy, GENE then computes the weight of each connection as the distance between the two connected neurons in the latent space. As the genome for $N$ neurons is only $dN$ parameters, the GENE encoding reduces the size of the genome by an order of magnitude for the same network architecture (see Table \ref{tab:genome_size}). With ES methods like \cmaes{} where the memory cost grows quadratically with the genome size, an 18-fold reduction of the genome means reducing the size of the covariance matrix in memory by 324 times, making large networks tractable.

The ``distance'' term is here used loosely to describe a function that takes two points in a Euclidean space and returns a value in $\mathbb{R}$. A proper distance function in the mathematical sense would only create positive values, but not having negative weights in the network would not allow to have inhibition behaviors where the activation of a neuron reduces the activation of another. GENE is defined with the pL2 distance function (\ref{eq:pL2}) based on the Euclidean distance with an additional factor allowing for negative values (\ref{eq:bounded_signed}).

\begin{equation}\label{eq:bounded_signed}
\alpha: \begin{cases}
 & \text{ if } x \geq 1: \alpha(x) = 1\\ 
 & \text{ if } x \leq -1: \alpha(x) = -1\\ 
 & \text{ else: }  \alpha(x) = x
\end{cases}
\end{equation}

\begin{equation}\label{eq:pL2}
  d_{pL2}(n_1, n_2) = \alpha( \prod_{i=1}^D n_1^i - n_2^i) \sqrt{\sum_{j=1}^D \left( n_1^j - n_2^j \right)^2 }
\end{equation}

$d_{pL2}(n_1, n_2)$ is then used as the weight of the connection from neuron $n_1$ to neuron $n_2$, and the ES optimizes the values of $n_1^i$ and $n_2^i$ for all latent dimensions $i$. Instead of hand-crafting distance functions like pl2 with human bias, we here propose to automatically discover them with meta-evolution. 

\begin{table}
    \centering
    \begin{tabular}{l|cccc}
         & \halfcheetah & \walker &  \hopper &  \swimmer \\
        \hline
        Direct & 19718 & 19590 & 18435 & 17922\\
        GENE   & 1102 & 1099 & 1069 & 1056\\
        \hline
        Ratio & 18 & 18 & 17 & 17 \\
    \end{tabular}
    \caption{Genome size for GENE ($d=3$) and direct encoding with the same neural network architecture used here, and the ratio between.}
    \label{tab:genome_size}
\end{table}

\subsection{Meta-evolution}
\label{sec:rw:meta_evo}

While usually manually defined through human engineering, algorithms can be automatically improved with the approach of meta-learning. 
Meta-learning methods have been applied to a variety of problems with the same underlying goal of replacing hand-crafted portions of a learning system with automatically discovered settings. This can be hyperparameters, as in Neural Architecture Search (NAS) and Hyperparameter Optimization, or entire portions of a learning algorithm. Learning to learn effectively is what drives the use of meat-learning methods. Evolutionary methods have a long history of being used in designing programs due to their derivative-free properties allowing them to work effectively on problems with unknown functional form \cite{schmidhuberEvolutionaryPrinciplesSelfreferential1987}. Their population-based nature allows them to leverage the latest hardware advances, making them ideal for hardware accelerators that utilize parallel operations.
Meta-learning works by turning parts of a learning algorithm into optimizable sections which are then learned by an outerloop.

Deep Reinforcement Learning methods have been a target for meta-learning to either automatically rediscover and improve on existing methods by representing algorithmic components as graphs, which can be optimized with Genetic Programming \cite{co-reyes_evolving_2021}. Meta-evolved functions can also be represented with neural networks optimized by an Evolution Strategy \cite{lu2022discovered}, with additional analysis being required to transform the learned network into a function that can easily be implemented as a standard learning algorithm. Gradient-free evolutionary meta-evolution methods \cite{jackson_discovering_2023} outperform gradient-based ones \cite{oh2020discovering} by allowing the meta-evolution to account for the complete training process of the optimized algorithm in one fitness value.

Meta-learning parts of evolutionary optimisation algorithms was also studied with Learned Evolution Strategies (LES, \cite{lange_discovering_2022}) and Learned Genetic Algorithms (LGA, \cite{lange_discovering_2023}) where a meta-learning loop with a self-attention based framework searches over the learning rates and weights assignment rules of an ES, or over the selection and mutation rate adaptions of a GA. They create fine tuned update rules which outperform previous human-engineered ES and GA respectively.

\subsection{Cartesian Genetic Programming}
As seen in previous work on the meta-evolution of learning methods (see section \ref{sec:rw:meta_evo}), the function to optimize can be represented by a neural network but the learned output then needs additional processing to be implemented as a standard learning method or studied analytically. Cartesian Genetic Programming (CGP, \cite{miller_cartesian_2008}) instead represents programs as graphs with discrete functions linked together, which can then be translated into interpretable code. CGP indexes the nodes of the GP graph with Cartesian coordinates and has been succesfully applied to a range of problems from boolean functions \cite{miller_cartesian_2008} to policy search in image-based Atari environments \cite{wilson_evolving_2018}.

\section{Method}
\label{sec:method}

\begin{figure*}[t]
    \captionsetup{justification=raggedright, singlelinecheck=false}
    \centering
    \includegraphics[width=0.8\textwidth]{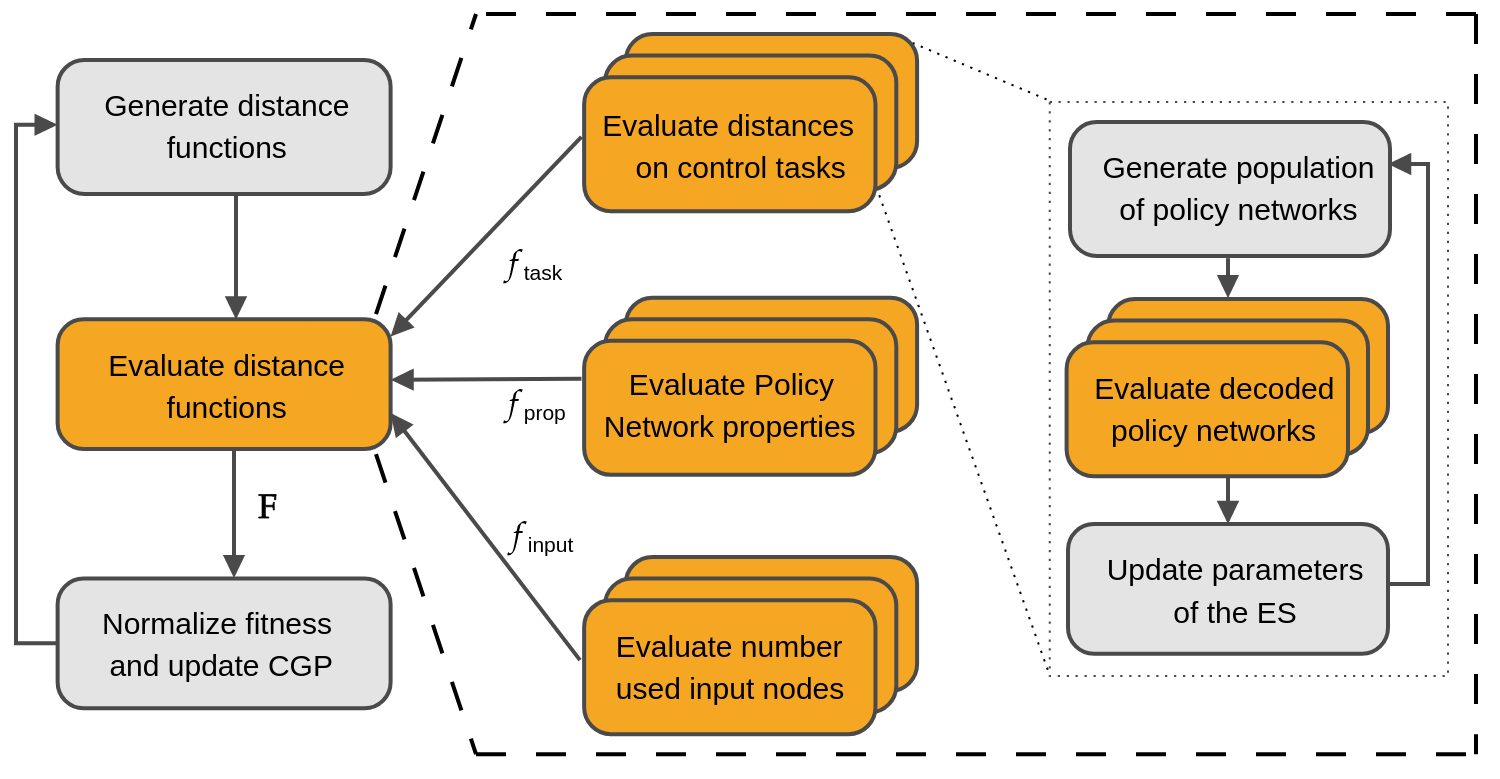}
    \caption{Meta-evolution process, depicting the different steps happening. In orange are the evaluation steps conducted in parallel. \textbf{Left:} The outer loop is a CGP process, generating distance functions and evaluating them. \textbf{Center}: The components of distance function evaluation. The final fitness used for the CGP outer-loop is $F=\beta f_\text{task} - (1-\beta)f_\text{prop} + \alpha f_\text{input}$. \textbf{Right:} The distance function being evaluated is used as GENE encoding in a loop to optimize neural networks with and ES.  This operation is done on both \halfcheetah\ and \walker\ with $f_\text{task}$ the sum of normalized final fitness scores.}

    \label{fig:meta_evol}
\end{figure*}

The pL2 distance function introduced in the GENE article has a number of shortcomings, including the impact of any mutation on both incoming and outgoing connections which can lead to a loss of expressiveness by the target network. However finding a better distance function is a complex process if done manually. Instead of defining a new function by hand, we use meta-evolution to learn a new one that performs well on a set of defined training tasks and evaluation criteria. The meta-evolution loop is defined in Fig. \ref{fig:meta_evol}.

\subsection{Distance function optimization}
Distance functions are represented and optimized using Cartesian Genetic Programming: the loop generates a population of candidate distance functions that are evaluated on a set of defined metrics then combined into the final fitness, which is used to update the state of the GP method. 

A GENE distance function, such as L2 or pL2, can be described by an acyclic directed graph taking as input the position vectors of the input neurons, e.g. their coordinates in the latent space introduced by the encoding. If the dimension of the latent space is $d=3$ then the input vector has size $2d = 6$. To allow for constants, CGP extends the input vector with additional nodes representing possible constants (see Appendix \ref{sec:appedix:cgp-constants}). The CGP graph has a single output, which is the weight of the connection in the network.
The resulting graph can either be used in its genome form, which is what we do during the meta-evolution process, or in its symbolic form after selection and extraction of a function and then used as is in the code thanks to the explainability of Genetic Programming.

The main meta-evolution loop is simply a standard Cartesian Genetic Programming algorithm optimizing the distance functions. 
Once the distance functions have been generated, they will be evaluated according to two criteria: training performance and generated network properties.

\subsection{Training performance}

Each distance is first evaluated on its ability to allow an ES to optimise neural networks as policies for control tasks over a full optimization run. 
During the evaluation phase, the distance functions are used to train policies on the \halfcheetah\ and \walker\ control tasks separately and the fitness of the best individual found is kept. Final maximum fitness values from \halfcheetah\ and \walker\ then go through min-max normalization to have comparable scales and are summed to give a $f_\text{task}$ score:
$$f_\text{task} = f_\text{halfcheetah} + f_\text{walker2d}$$

We use two environments for the training phase of the meta-evolution to avoid overfitting the encoding on one problem. While this is expensive, the meta-evolution loop is only needed once to find a good distance function which can then be applied to any problem.

\subsection{Evaluating network properties}
The L2 distance function first tested for GENE is a good example of ill-formed distance function as it can only create positive values, removing the possibility for inhibition behavior to emerge in the network. To bias the meta-evolution search towards functions that can create usable neural networks we introduce regularization terms in the CGP fitness, described below.

\subsubsection{Weight distribution}
Neural networks trained with direct encoding show a weight distribution following a Gaussian distribution centered around $0$ (see Fig. \ref{fig:weight_distrib}), which is consistent with standard ANN initialization methods as it allows for a balanced number of activator and inhibitor connections  \cite{hanin2018start}. We favor distance functions which will create networks with this property by sampling random genomes from a Gaussian distribution and measuring the distribution of the weights.
We note the value of the mean of the weights as $f_\text{mean}$ as well as the square of the difference between the standard deviation and $0.5$ as $f_\text{std}$.
These penalties encourage networks whose weight distribution is that of a Gaussian centered at $0$ and with a standard deviation of $0.5$.

\subsubsection{Input distribution restoration}

Neural networks trained on data should be able to pick out the underlying distribution of the data. In order to do so, the initial weight distribution should ideally be symmetric \cite{hanin2018start}. This property is tested by running samples throughout the network and evaluating the obtained distributions mean with respect to the input one. The absolute difference from the mean is denoted as $f_\text{sym}$.

\subsection{Using all inputs}
CGP has a bias towards functions which do not use all inputs due to the graph representation and mutation operator. To push the evolved distance function to use all the dimensions of the projected space, we introduce a bonus term $f_\text{input}$ scaled by an hyperparameter $\alpha$ (see Fig. \ref{fig:meta_evol}) which is maximized when all 6 input nodes of the graph are used. Unused dimensions in a learned distance functions will hence highlight that no evolutionary advantage was found by increasing the graph complexity.

\subsection{Weighted integration}

To calculate the final fitness used for a distance function, we first sum the network weight distribution terms to make $f_\text{prop}$:

$$f_\text{prop} = f_\text{mean}+f_\text{std}+f_\text{sym}$$

This term expresses how far the evolved network weights are from the desired distribution, i.e. that the network weights are distributed normally around 0 with a small standard deviation and represent a symmetric function. As this term is expressed as a penalty, the term $f_\text{prop}$ is subtracted from the task fitness $f_\text{task}$ in the final fitness $F$:

$$F=\beta f_\text{task} - (1-\beta)f_\text{prop} + \alpha f_\text{input}$$

The fitness $F$ is therefore the sum of both task fitness values and the bonus for using all inputs $f_\text{input}$, minus the penalty of network properties. We used a $\beta$ of $1/3$, meaning that greater focus was given to the network properties than the task fitness.

\section{Experiments and Results}
\label{sec:results}

\subsection{Discovered GENE encodings}
	
\definecolor{Gray}{gray}{0.9}
\begin{table*}[t]
    \centering
    \begin{tabular}{c|l|cccc}
        ID & Function & \halfcheetah & \walker &  \hopper &  \swimmer \\
        \hline
        Direct &  &       1561 &      \textbf{2694} &    1468 &      297 \\
        pL2    &   &      1501 &      \underline{2172} &    1754 &      \underline{298} \\
        \hline
        LD-10 & $x_2 $ &         7649 &      1194 &     232 &      214 \\
        LD-79  & $sin(exp(z_1) * y_2) $ &         7793 &      1326 &     642 &      253 \\
        LD-204 & $sin(((((z_2 - x_1) > z_2) + (z_2 - ((z_2 - x_1) > 0.1))) < abs(z_2)) * y_2) $ &         7669 &      1574 &    \underline{2164} &      286 \\
        LD-206 & $sin(abs(z_2 > x_1) * y_2) $ &         7853 &      1298 &    2040 &      193 \\
        LD-318 & $abs(z_2 > x_1) * y_2 $ &         \underline{7898} &      1334 &    1877 &      243 \\
        LD-352 & $\sqrt{z_2 > x_1} * y_2 $ &         \underline{7898} &      1334 &    1877 &      243 \\
        \rowcolor{Gray} \textbf{LD-367} & $\sqrt{x_2 > z_1} * y_2 $ &         7766 &      1944 &    \textbf{2211} &      \textbf{291} \\
        LD-376 & $((x_2 > x_1) / 1) * y_2 $ &         7744 &      1650 &    1850 &      250 \\
        LD-573 & $(((1 * z_2) < ((x_2 - (1 * x_1)) - cos(x_2))) < (0.1 * ((1 < abs(z_1)) < z_2))) $ &         7701 &      1196 &     629 &      167 \\
        LD-626 & $sin(x_2 < x_1) * y_2 $ &         \textbf{7910} &      1241 &    1046 &      233 \\
         \hline
    \end{tabular}
    \caption{Top 10 distance functions found by the meta-evolution with corresponding average final scores on control tasks. Highest fitness in bold, second highest underlined. The best distance function LD-367 is used in the rest of the paper.}
    \label{tab:top_dist}
\end{table*}

During the meta-evolution process, the distance functions that obtain the highest fitness are archived. From these archived distance functions, we evaluate the best $10$ on a series of environments, including new environments \hopper{} and \swimmer{} not seen during training to assess their generalization capabilities. Functions and results are presented in Table \ref{tab:top_dist}. The ID corresponds to the generation in the meta-evolution where the distance appeared, with "LD" standing for "Learned Distance". We use LD-367 as learned distance in the other figures as it performs well on all environments. Functions are shown as learned by CGP, but they can be reduced as the square root or the absolute value of a boolean does not change its value. LD-318 and LD-352 are for example functionally identical and reach the same scores.

\subsection{Best learned function}
\label{subsec:best-df}

\subsubsection{Function graph}
\begin{figure}
    \captionsetup{justification=raggedright, singlelinecheck=false}
    \centering
    \includegraphics[width=0.25\textwidth]{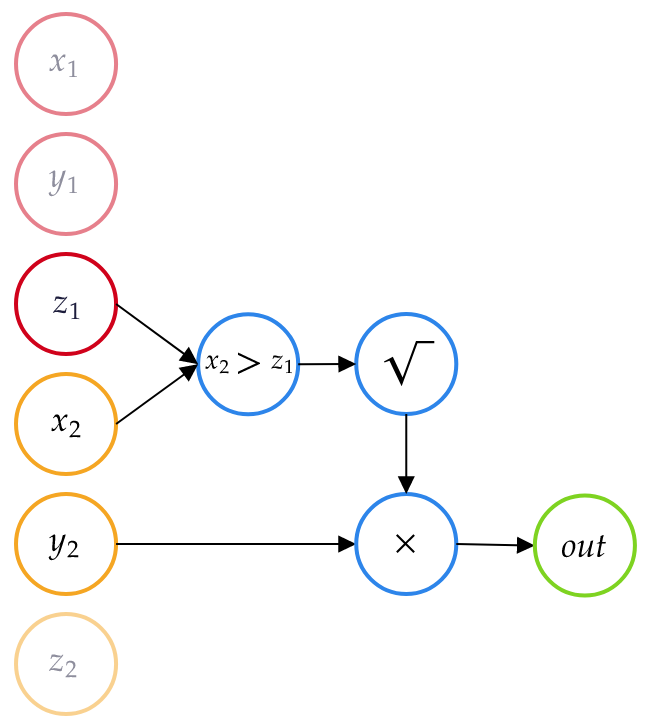}
    \caption{Visualization of the CGP graph of the learned distance function. In \textbf{red} are the input coordinates of the first neuron and in \textbf{orange} are the input values of the second neuron. In \textbf{blue} are the operator nodes, used by CGP to construct a function. The output node is depicted in \textbf{green}. The function can be simplified as $d=(x_2>z_1)y_2$, as $>$ is a boolean operator. In total, 3 input nodes, 3 operator nodes and the single output node are used.}
    \label{fig:learned-df-367}
\end{figure}

We select LD-367 as the best distance function, reaching high fitness values on all environments, including ones not seen during the meta-evolution. Its graph is can be visualized in Fig. \ref{fig:learned-df-367}.

The function is very simple, using only 3 of the 6 input coordinates. With $n^{l}_1$ and $n^{(l+1)}_2$ two nodes connected by the weight $w^l_{1, 2}$, their respective position vectors in the latent space are $n^{l}_1 =[x_1, y_1, z_1]$ and $n^{l+1}_2 =[x_2, y_2, z_2]$.
The value of this specific weight is defined by the decoding operation in equation \ref{eq:df367}.
\begin{align}\label{eq:df367}
    w^l_{1, 2} & = d_{367}(n^{l}_1, n^{l+1}_2) \\
    & = (x_2 > z_1)y_2    
\end{align}   

\begin{figure}
    \centering
    \includegraphics[width=0.5\textwidth]{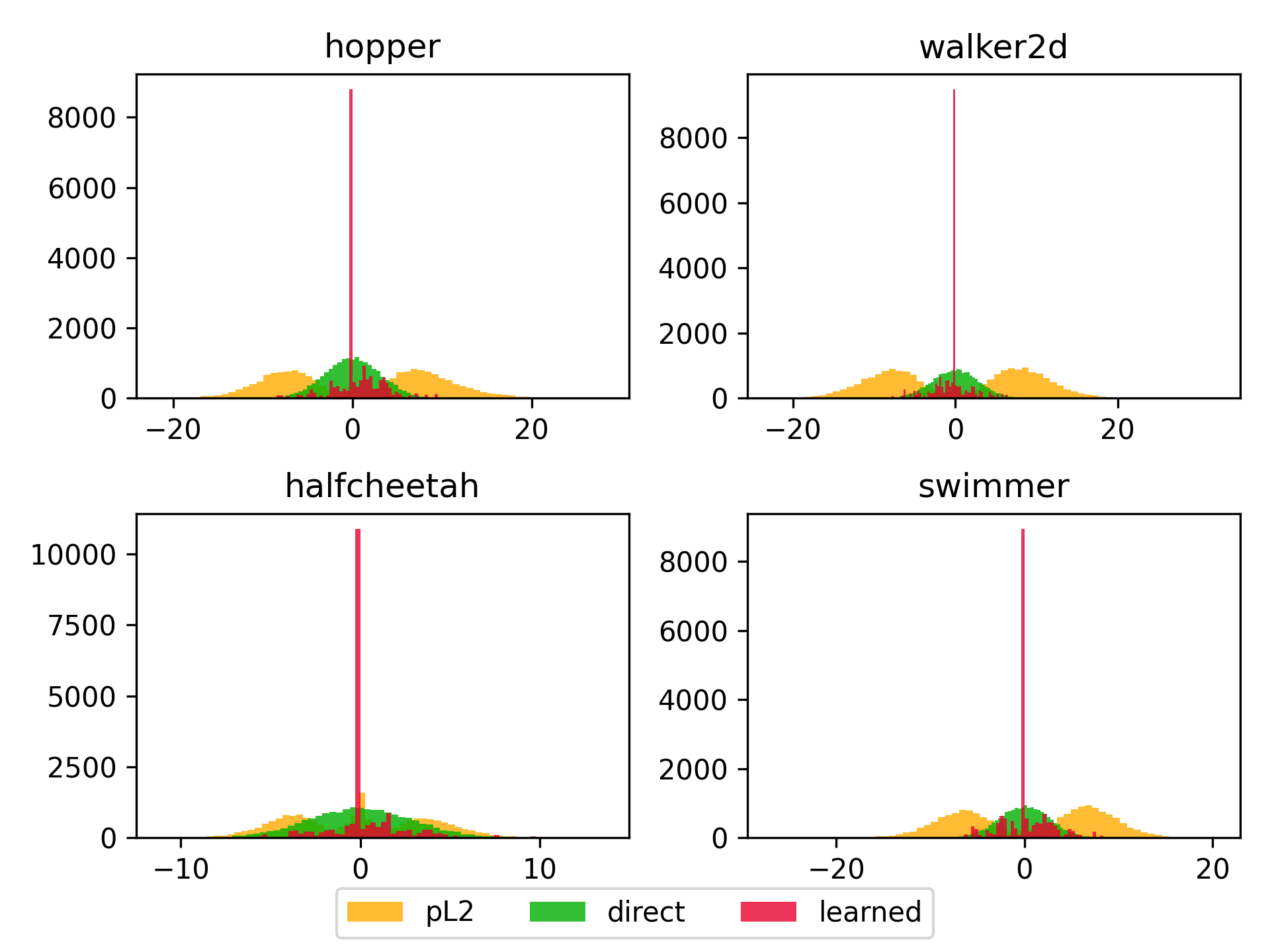}
    \caption{Weight values distribution for neural networks trained with direct encoding and GENE with pl2 and learned distances. }
    \label{fig:weight_distrib}
\end{figure}

As $>$ is a boolean operator we get $w^l_{1, 2} \in \{0, y_2\}$, so this learned distance function acts a pruning operator, putting some weights to $0$ and propagating $y_2$ to the others (connection-specific).
Of the 64 available operator nodes that CGP could use to construct a distance function, only 3 were ultimately used. But these 3 nodes, in addition to GENE, were enough to simulate a pruning operator and ultimately beat a carefully hand-crafted distance function such as pL2 or tag-gene\cite{templier_geometric_2021}. We show in Fig. \ref{fig:weight_distrib} the distribution of weight values in neural networks trained with each encoding: the learned encoding

\subsubsection{Evaluating the encoding}
\begin{figure}
    \centering
    \includegraphics[width=0.45\textwidth]{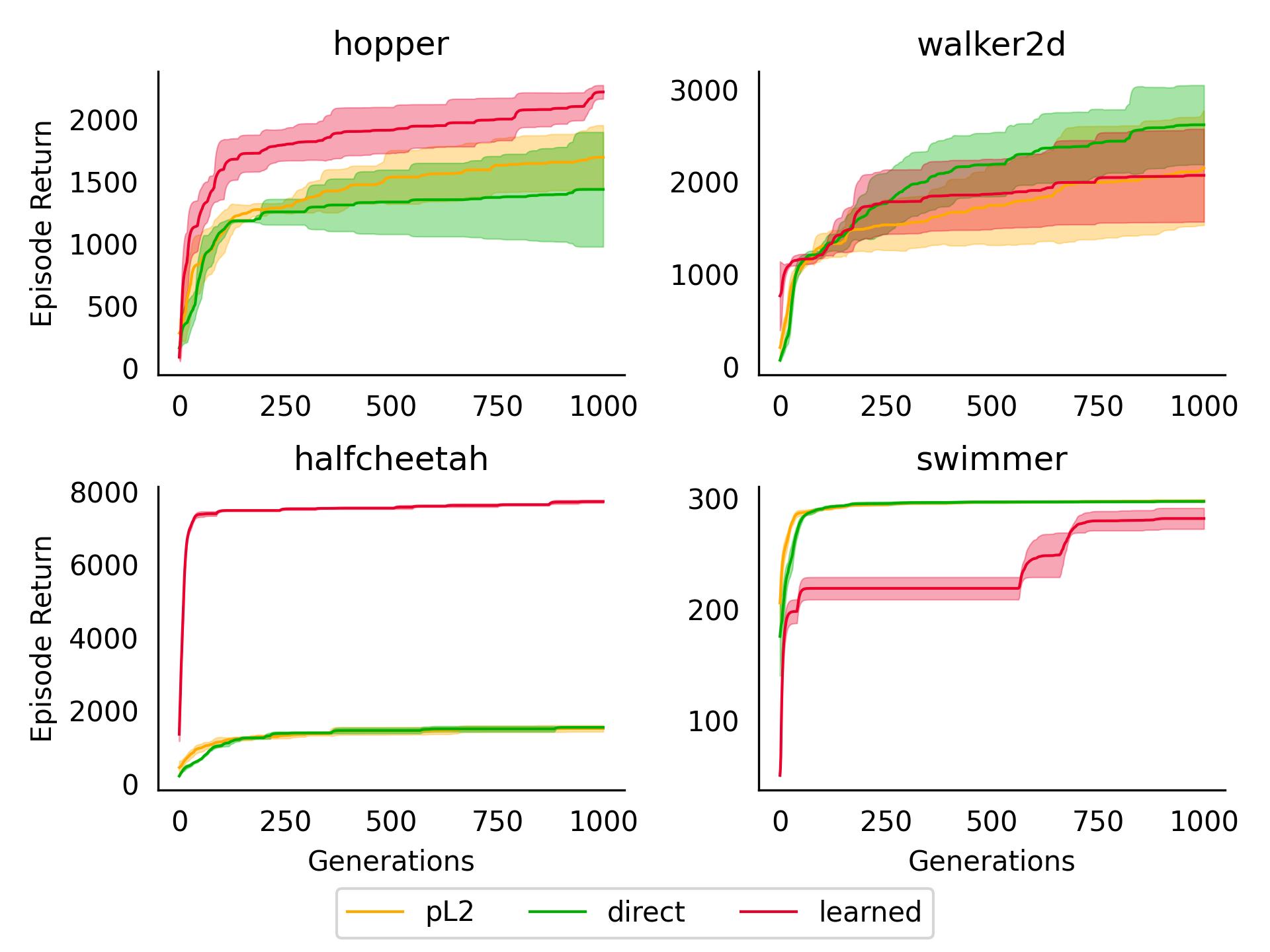}
    \caption{Maximum fitness reached by an ES with each encoding, averaged over $5$ runs.}
    \label{fig:top_0_fit}
\end{figure}

The learned encoding is then compared to direct encoding and to the hand-crafted original pl2 distance for GENE on control tasks \halfcheetah{}, \walker{}, \swimmer{} and \hopper{}. Training curves are presented in Fig. \ref{fig:top_0_fit}. 

On 3 of the 4 environments, the learned distance functions performs better than pL2 and on 2 of them, \halfcheetah{} and \hopper{}, it even beats direct encoding.
On \swimmer{}, the learned distance function takes longer to attain a good return value, with a very small gap between pL2 and direct encoding. By extrapolating the curve, we can imagine that the learned distance function has still plenty of room for improvement in comparison to pL2 and direct encoding who converge really fast but are stuck for the rest of the optimization. 

\begin{figure}
    \centering
    \includegraphics[width=0.5\textwidth]{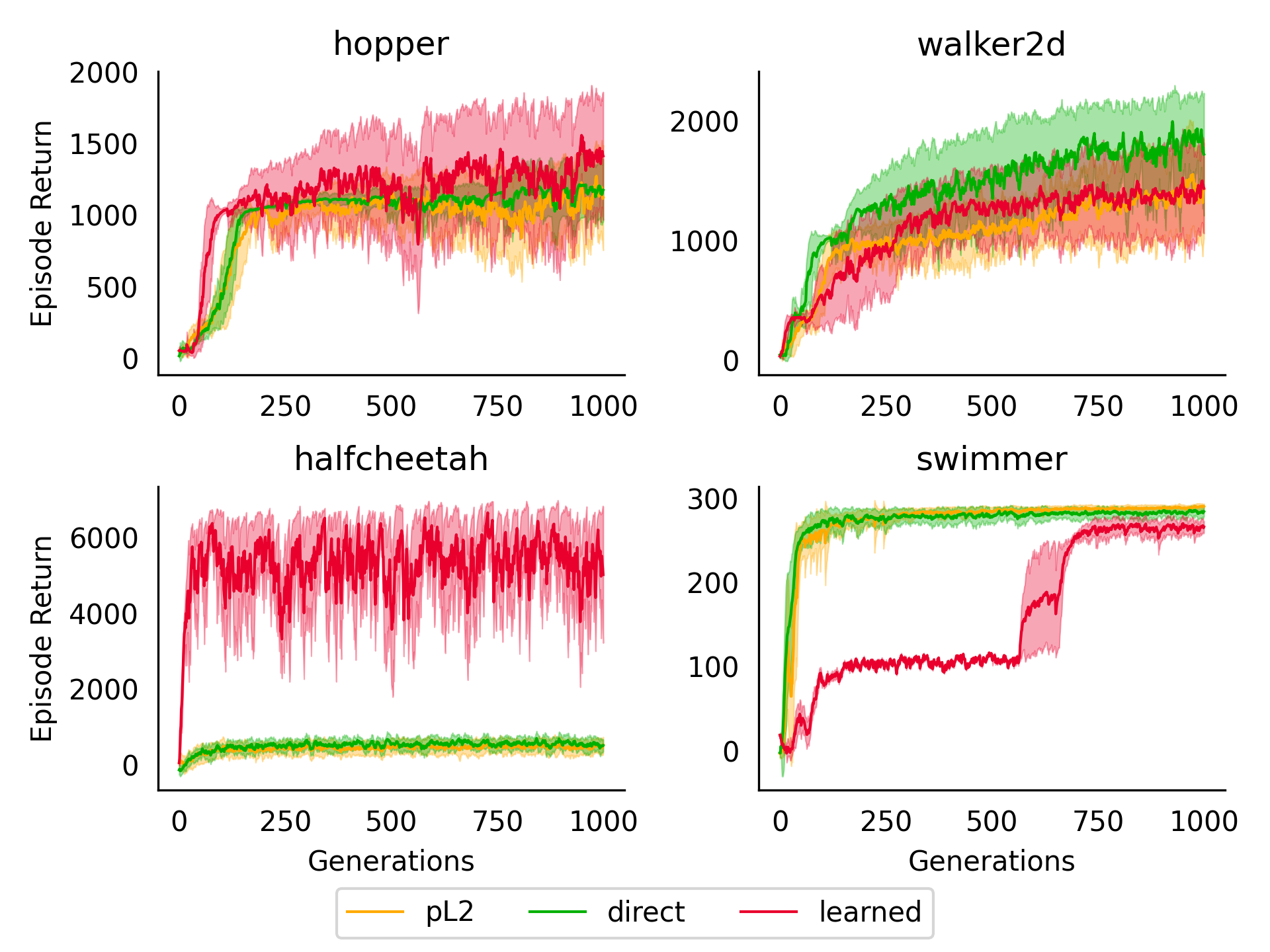}
    \caption{Fitness of the center individual of the population, averaged over $5$ run. The noisiness can be explained by the fact that the mean of the population is not very stable compared to the best observed individual. }
    \label{fig:mean_fit}
\end{figure}

Since Evolutionary Strategies have a high variance between individuals of the same population, and even if what is ultimately of interest is the best found individual, a more robust measure of the performance of an ES is to measure the centre of the population. The same center used as a reference point to sample new candidates. Fig. \ref{fig:mean_fit} show us the learning curves obtained by evaluating the center of the population. The obtained results lead us to draw the same conclusion than those for the overall best individual.

\subsection{Meta-evolution}

\begin{figure}
    \centering
    \includegraphics[width=0.48\textwidth]{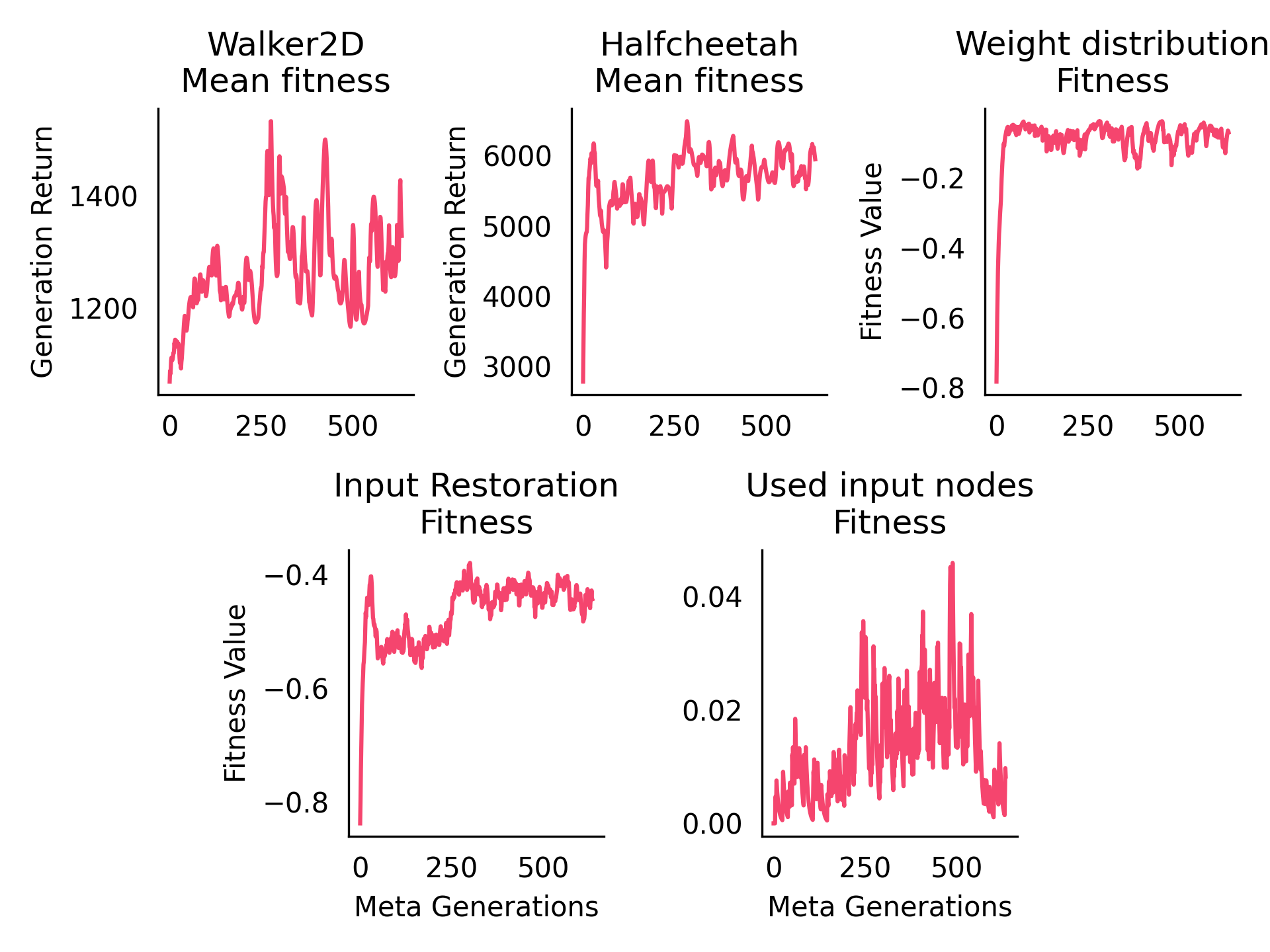}
    \caption{Training curves of the Meta-evolution, for a few collected fitness values during training.}
    \label{fig:meta-evol-graph}
\end{figure}

Taking a step back from the final result, the whole meta-evolution process can be analysed in Fig. \ref{fig:meta-evol-graph}. The process ran for 635 generation and since no notable improvement can be observed after around 500 generations, we decided to stop the training.
After quickly reaching 1200, the fitness value for \walker{} oscillates between 1200 and 1400, while showing a slight upward trend. The training on \halfcheetah{} quickly arrives at very good-performing policies using the learned distance functions.

As expected, the 2 enforced network properties (weight distribution and input restoration capabilities of the policy networks) show fast convergence. 
So the learned distance functions correctly enforced the defined network properties and used this to their advantage to find good distance functions.
The used input nodes property, however, had no defining influence on the learning process.

The proportion of distance functions that use all 6 input values, i.e. all information in the coordinates of the projected neurons, is very low throughout training. This suggests that using all input values is not mandatory to find good performing distance functions in the current setting.

\subsection{Analysing network properties and fitness}
In order to observe the evolution of the characteristics of the political networks, we also evaluated these properties during the evaluation of the distance functions found, separating them by encoding type used.

\begin{figure}[ht]
    \centering
    \includegraphics[width=0.5\textwidth]{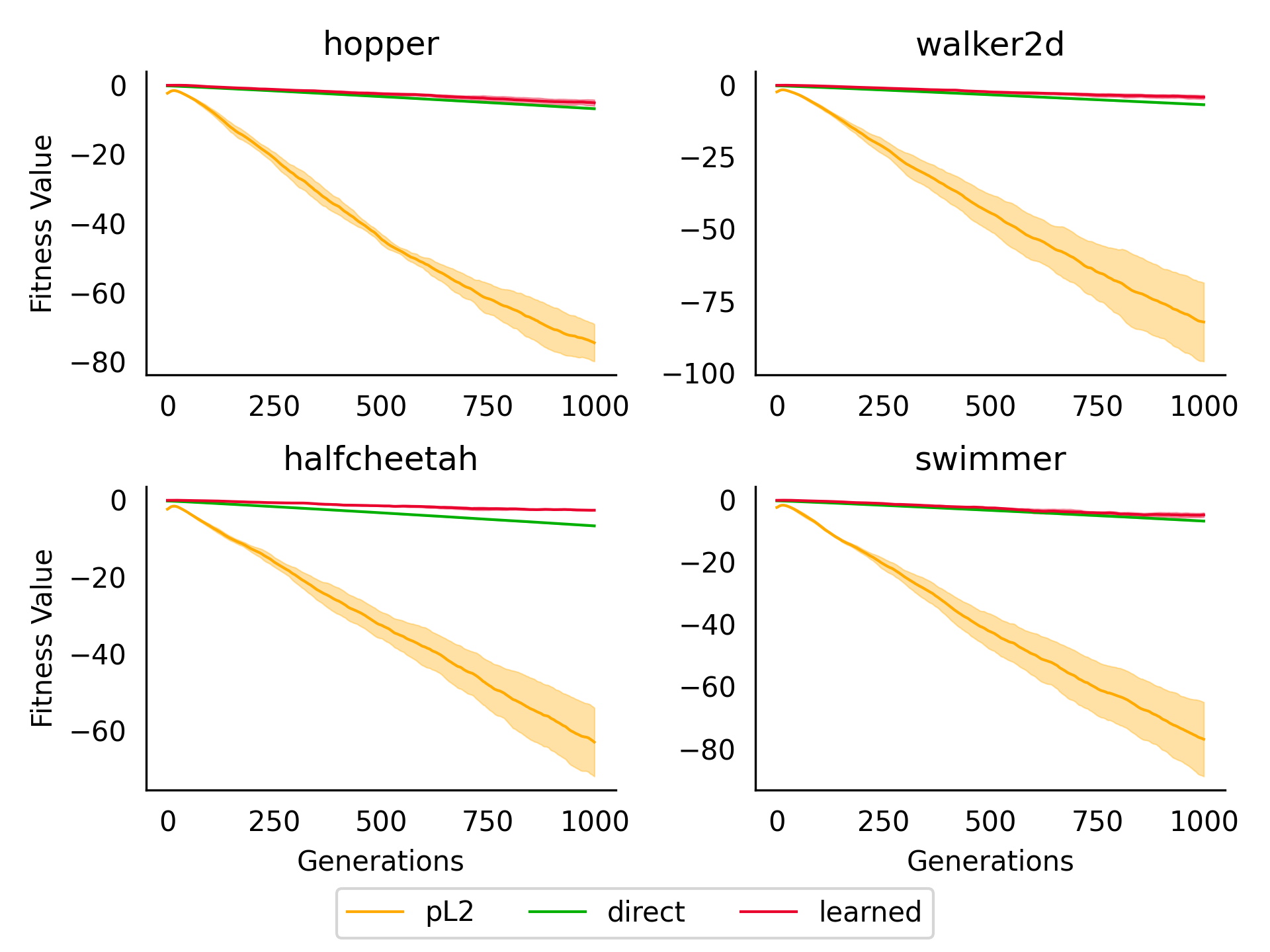}
    \caption{Fitness of the weight distribution of the policy networks decoded with the learned distance function over the generations. Values are averaged during training over all polices and over $5$ different runs.}
    \label{fig:f_weight_distribution}
\end{figure}

We note in Fig. \ref{fig:f_weight_distribution} that the distribution of the policy network weights encoded with the learned distance function (see \ref{fig:learned-df-367}), remains well centred around $0$ and with a variance close to $0.5$. This can be explained by the low penalty value obtained during training. 

What is surprising in Fig. \ref{fig:f_weight_distribution} is the high number of weights with the exact value of 0. When analyzing the evolved distance functions, we note that many of them include logic which results in exactly 0 for weights, i.e. the use of logical functions like $(x_2 > z_1)$. This results in sparse networks: networks which have a small percentage of non-zero weights. Sparse neural networks have been well-studied \cite{hoefler2021sparsity} and it is known that sparse versions of dense networks can be found through pruning \cite{frankle2018lottery}, however it is surprising that evolution discovered a way to make sparse networks through the evolved distance functions, and that these distance functions generalize well to other tasks. This joins the analysis of \cite{lange_lottery_2023}, that high-performing sparse networks can be found by evolutionary strategies, while also proposing an encoding scheme to search only over sparse networks.

\begin{figure}[ht]
    \centering
    \includegraphics[width=0.5\textwidth]{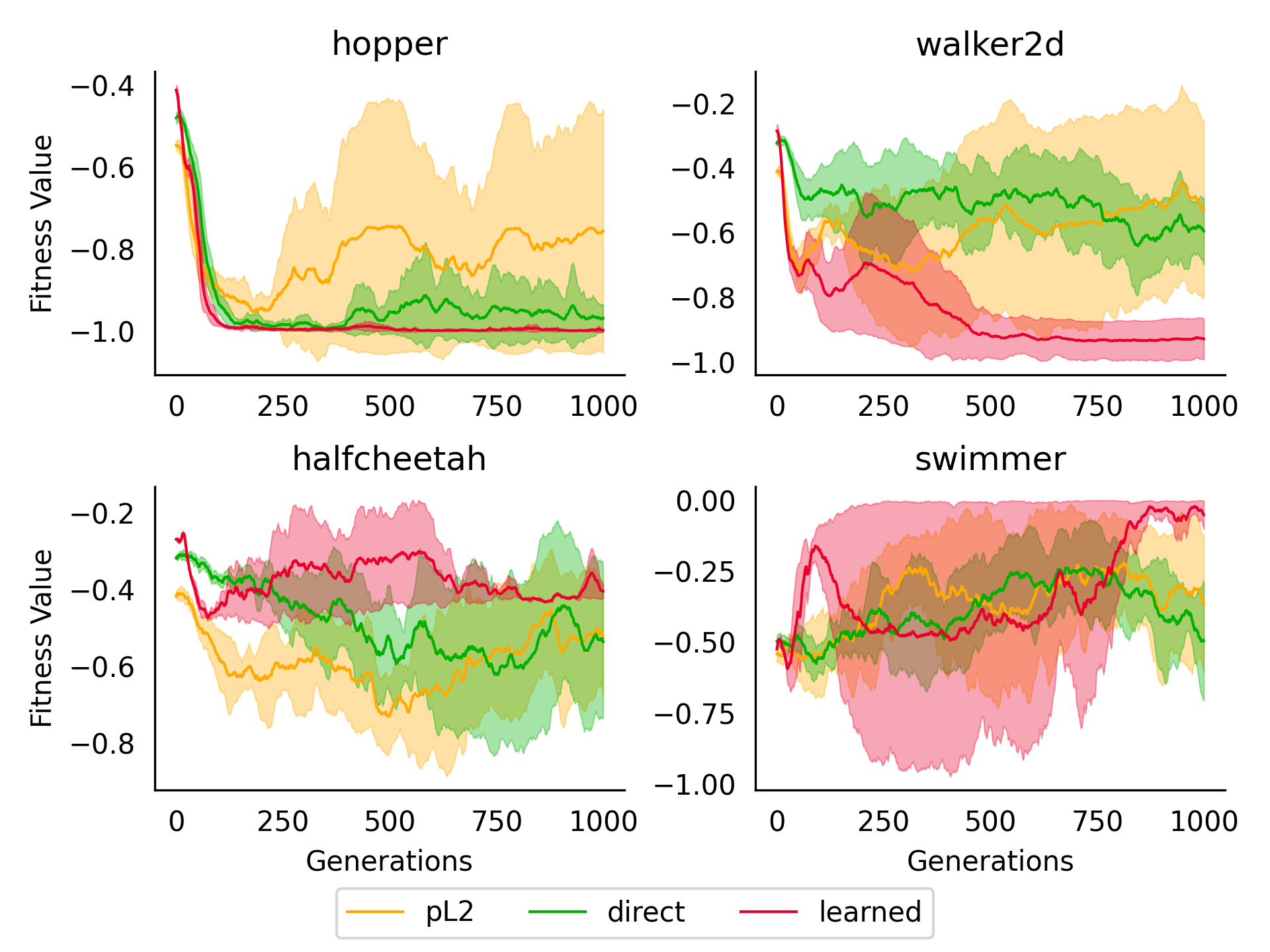}
    \caption{Fitness of the input restoration ability of the policy networks decoded with the learned distance function over the generations. Values are averaged during training over all polices and over $5$ different runs.}
    \label{fig:f_input_restoration}
\end{figure}

The importance of the input distribution restoration property is harder to analyze, see Fig. \ref{fig:f_input_restoration}. It seems that it is entirely task-specific and that there are no links between the best found policy, and the networks ability to enforce this specific property.
Indeed, the input distribution restoration property is only of interest as an initial starting point for an untrained neural network.

\section{Discussion}
\label{sec:conclusion}

We showed that meta-evolving a distance function for the GENE encoding allows for better results on a set of continuous control benchmarks. Leveraging CGP, we were able to obtain a learned distance function that is small, interpretable and thus easily portable.
Throughout the different experiments we argue that the selected network properties have a positive impact on the learning process. Surprisingly, multiple distance functions evolved to encode sparse networks that, despite their sparsity, perform well across environments.

The main limitation of this work is the computational cost of the meta-evolution. Evolving a specific distance function for a specific set of problems is hardly tractable. This motivated the use of CGP with the hopes that the output could be reuse and generalize to unseen problems. As we demonstrate with the hopper and swimmer environments, the evolved distance functions do generalize to new tasks.

A direction for future work would be to define more neural network properties and carefully evaluate them to select those that are truly important for a network policy. This could guide the meta-evolution process more effectively and discover even more interesting and better performing distance functions. 
Furthermore, we used the same tasks and network architectures for all evaluated distance functions in this work. In future work, we aim to use curriculum learning to increase the complexity of tasks and the number of different architectures tested as the distance function improves over evolution.

This work demonstrates that evolution can be a powerful tool to learn new neural network representations. By using an interpretable form of evolution, genetic programming, we were able to study the evolved representation schemes and understand their functioning. Further study could give rise to new, evolved types of neural networks, such as the sparse, compressed encoding discovered in this work.

\nopagebreak
\bibliography{references}
\nopagebreak
\appendix
\section{Appendix}
\label{sec:appendix}

\subsection{Implementation details}
All the evaluations of the external and internal loops are carried out in parallel using GPU hardware acceleration with JAX\cite{jax2018github} and open-source libraries such as Evosax\cite{evosax2022github} and Brax\cite{brax2021github} that allow for the environment rollouts to run on the GPU. Research code will be made available on github. 
The search space of our CGP meta-evolution loop is defined by the number of nodes used, in our case $64$ to guarantee a certain expressiveness, and by the size of the set of pre-defined $\mathcal{F}$ operators. 

\subsection{Hyper-parameter table}
\begin{table}[h]
    \centering
    \begin{tabular}{ l | r }
        \hline
        \textbf{CGP Hyperparameter} & \textbf{Value}  \\ 
        \hline
         Meta-population size & 32 \\
         Number of generations & 5000 \\
         Number of nodes & 64 \\
         Replacement for NaN values & 0 \\
         Number of used constants & 2 \\
         number of used functions & 12 \\
         Probability of mutating a function & 0.15 \\
         Probability of mutating an input & 0.15 \\
         
         \hline \hline
         \textbf{Inner-loop Hyperparameter} & \textbf{Value}  \\ \hline
         Inner-loop policy networks population size & 32 \\
         GENE encoding dimension $d$ & 3 \\
         $\beta$ & ${1/3}$ \\
         $\alpha$ & ${4\beta}$ \\
         Activation functions used for the policies & $\tanh$\\ 
         Strategy used for policy search & Sep CMA-ES \\
         Policy network internal layer architecture & [128, 128] \\
         Number of generations for \halfcheetah & 500 \\
         Number of simulation steps for \halfcheetah & 1000 \\
         Number of generations for \walker & 1500 \\
         Number of simulation steps for \walker & 1000 \\
         \hline
    \end{tabular}
    \label{tab:hyper-parameters}
\end{table}

\subsection{Cartesian Genetic Programming}
\label{sec:appedix:cgp-constants}
Only $2$ constants are used as possible input nodes to be selected by CGP, $0.1$ and $1$. This creates an input vector of size $8$, not shown in figure \ref{fig:learned-df-367}.

Operators used are 

\begin{table}[h]
    \captionsetup{justification=raggedright, singlelinecheck=false}
    \centering
    \begin{tabular}{ | c | c | }
        \hline
         \textbf{Operator} & \textbf{Operands}  \\ 
        \hline
         $+$ & 2 \\
         $-$ & 2 \\
         $*$ & 2 \\
         $/^*$ & 2 \\
         $|\ |$ & 1 \\
         $\exp$ & 1 \\
         $\sin$ & 1 \\
         $\cos$ & 1 \\
         $\log$ & 1 \\
         $\sqrt{\ }^*$ & 1 \\
         $<$ & 2 \\
         $>$ & 2 \\
         \hline
    \end{tabular}
    \caption{Started operators are protected by being passed through the absolute function, i.e. to avoid impossible operations. Operators that accept only $1$ operand can only accept $1$ incomming link in the CGP graph.}
    \label{tab:cgp_operators}
\end{table}

\end{document}